\documentclass[journal]{IEEEtran}

\usepackage{amsmath}
\usepackage{array}
\usepackage[utf8]{inputenc}
\usepackage[margin=1in]{geometry}
\usepackage{subcaption}
\usepackage{xcolor}
\usepackage{graphicx}
\usepackage{comment}
\usepackage{gensymb}
\usepackage{verbatim}
\usepackage{url}
\begin{document}

\title{Grasping Benchmarks: Normalizing for Object Size \& Approximating Hand Workspaces}
%\title{Assessing Robot Hand Capabilities by Task}
\author{John~Morrow*, Nuha~Nishat*, Joshua Campbell*,  Ravi~Balasubramanian, and~Cindy~Grimm \thanks{morrowjo, nishatn, campbjos, ravi.balasubramanian, cindy.grimm at Oregon State University, *all authors contributed equally} }% <-this % stops a space

\maketitle

\begin{abstract}
The varied landscape of robotic hand designs 
makes it difficult to set a standard for how to measure hand size and to communicate the size of objects it can grasp. 
Defining consistent workspace measurements would greatly assist scientific communication in robotic grasping research because it would allow researchers to
1) quantitatively communicate an object's relative size to a hand's and 2) approximate a functional subspace of a hand's kinematic workspace in a human-readable way.
The goal of this paper is to specify a measurement procedure that quantitatively captures a hand's workspace size for both a precision and power grasp. 
This measurement procedure uses a {\em functional} approach --- based on a generic grasping scenario of a hypothetical object --- in order to make the procedure as generalizable and repeatable as possible, regardless of the actual hand design. 
This functional approach lets the measurer choose the exact finger configurations and contact points that satisfy the generic grasping scenario, while ensuring that the measurements are {\em functionally} comparable.
We demonstrate these functional measurements on seven hand configurations. Additional hand measurements and instructions are provided in a GitHub Repository.
\end{abstract}

\section{Introduction}

\begin{figure*}[ht]
    \centering
    \includegraphics[width=0.925\linewidth]{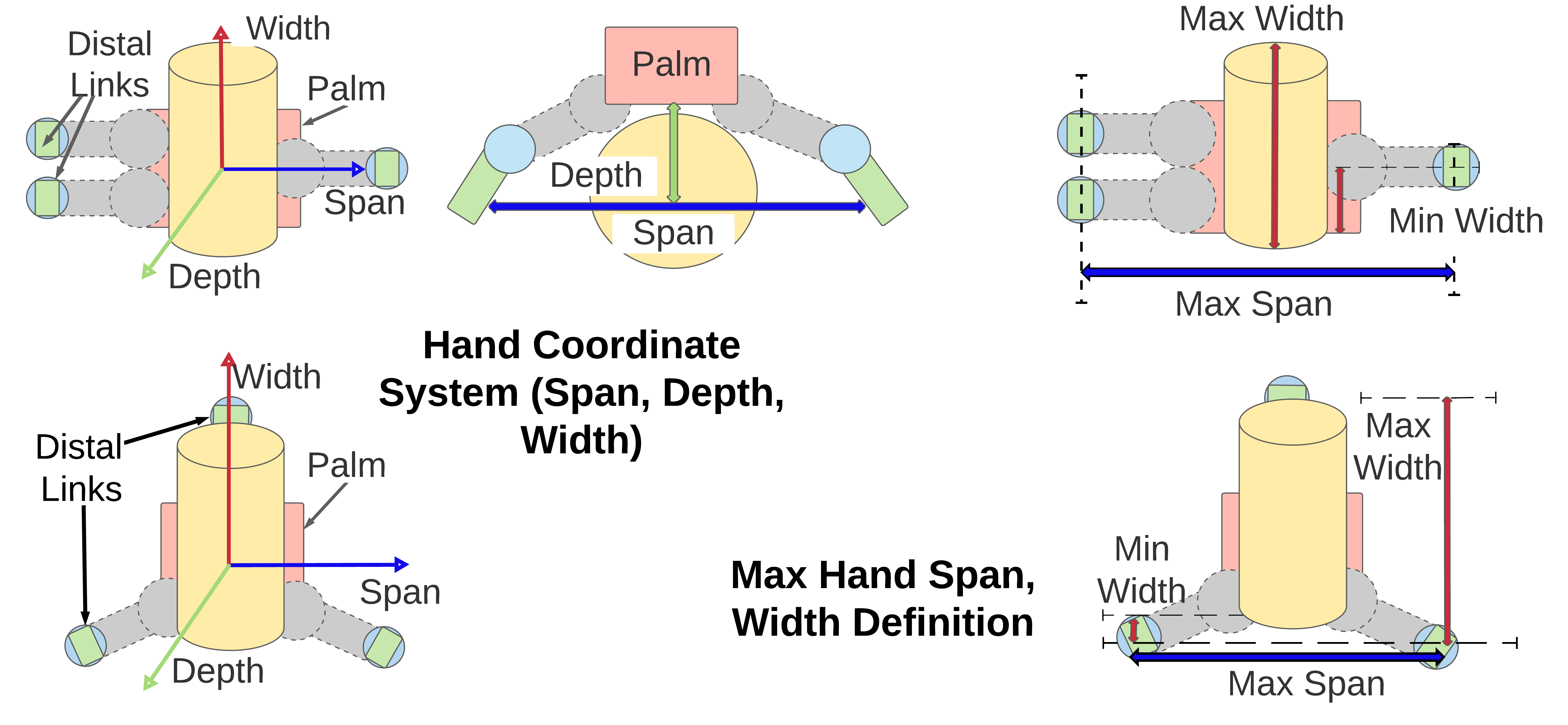}
    \caption{\textit{Upper left:} Defining the axes for a simple (top) and symmetrical (bottom) grasp. The fingers should be positioned as if the hand would grasp a hypothetical cylinder resting on a table. \textit{Right:} Defining the minimum and maximum Span and Width measurements.
    \vspace{-0.3cm}}
    \label{fig:axes}
\end{figure*}
%%%%%%%%%%%%%%%%%%%%%%%%%%%%%%%%%%%%%%%%%%%%%%%%%%%%

A common goal amongst grasping and manipulation researchers is to use universal benchmarks to characterize robot hand performance~\cite{bib:scientific-method,bib:guesteditorial} --- similar to the fields of Computer Vision and Graphics. However, unlike these fields, robot grasping and manipulation involve a hardware interaction between hands and objects%, the kinematics of the hands themselves)
. Hardware can vary substantially in terms of both kinematics and size, and the resulting physical interactions between the hand geometry and the object's during grasping and manipulation introduces even more variation. Establishing a vocabulary or methodology for characterizing these interactions is an on-going challenge in the field of robotic benchmarking. In this paper we propose a solution for one part of this problem: How to measure the functional workspace of a hand in order to normalize this space across different hands  --- i.e., what is a ``small'' versus a ``large'' object in the context of a specific grasp-type or manipulation task. 

Previously, researchers have addressed this problem by using standardized object sets (such as ~\cite{bib:ycb-objectset, bib:sensorized-objectset}) and simple task-based measures to characterize a robot hand's capabilities~\cite{bib:task-based}. These standards help to normalize the object shape and task, but do little to help with comparisons {\em across} hand designs --- if hand workspaces are different sizes, success or failure may simply be a result of the object's size relative to the hand, not the ability of the hand to perform the task. Even something as simple as characterizing in-hand manipulation of a cube presents challenges because there is no clear definition of a ``normal'' sized cube since it depends on the hand's overall size {\em and} manipulation space.

We take a step towards a standard characterization method by clearly defining the characteristics of two common grasps --- Precision and Power~\cite{bib:grasp-taxonomy} --- and how to approximate the graspable workspace. This addresses two specific problems: 1) How to quantitatively measure and compare an object's size in relation to a hand's size and 2) Measuring and communicating, in a human intuitive way, how a hand actuates and the range of object {\em volumes} it can grasp across its actuation space. 

Problem 1) arises with the simple task of defining a small, medium, or large object for a robot hand in a generalized way. Given a hand's size, a simple solution is to provide a formula for normalizing an object's size to the hand's (Section~\ref{sec:usability}
). Unfortunately, there is no consensus on what a hand's size {\em is}, nor is there a consensus on what ``small'' means. Simply measuring the hand in its open, rest state is problematic because some hands can't grasp {\em anything} in this configuration (most articulated fingers) while a parallel jaw gripper can. Our approach addresses this problem by defining a functional task (holding a canonical object) and performing the measurements based on that configuration.

Problem 2) relates to the field's lack of {\em functional} characterization methods. This is evident when buying robot hands. Often, the description only includes the dimensions of the hand, leaving determining the actuation space to the buyer. This does little to define its actuation and the range of object sizes it can actually grasp. Unfortunately, kinematic spaces can be complex, especially for underactuated hands. Completely characterizing the space, on the other hand, would create a confusing and complex representation. Our approach is to represent the most salient areas of the hand's kinematic workspace in a human-readable way. For this, we turn to the notion of {\em Power} and {\em Precision} grasps~\cite{bib:grasp-taxonomy}. We provide {\em functional} definitions of these grasps and descriptive (rather than prescriptive) methods for measurement. Although this approach leaves some room for interpretation on the part of the person doing the measuring, we feel that this flexibility is necessary in order to support characterization of the wide variety of available hand morphologies.

One final goal of our approach is to find a balance between conciseness and descriptiveness. We also aim to limit the total measuring time to half an hour. Our measurement approach is also applicable to CAD models with both actuated and unactuated implementations. This enables comparison between hand designs before adding a specific actuation mechanism. 

\textbf{Contributions: }
In summary, we propose a quantitative robotic hand workspace measurement that: a) can be applied to a wide variety of fingered hand designs, b) can be used to normalize an object's size to a hand's size for benchmarking purposes, and c) can be used to communicate a practical approximation of a hand's graspable workspace in a human-readable way. 

To demonstrate the approach we applied it to two finger configurations of the Yale Openhand Model O~\cite{bib:yale-openhand}, and one finger configuration of the Barrett hand, Yale Openhand T-42, Robotiq 2F-85, Kinova Jaco 2 3 finger, and the right hand of the first author. For sake of space, we only include the detailed measurements of the Model O configurations in the paper, although all size measurement sets will be discussed in Section \ref{sec:results}. The measurements for the five additional hands, along with complete measurement instructions, can be found at \url{https://github.com/OSUrobotics/Applied-Hand-Dimensions}~\cite{bib:github}.

\section{Related Works}
%%%%%%%%%%%%%%%%%%%%%%%%%%%%%%%%%%%%%%%%%%%%%%%%%%%%
\begin{figure*}
    \centering
    \includegraphics[width=0.95\textwidth]{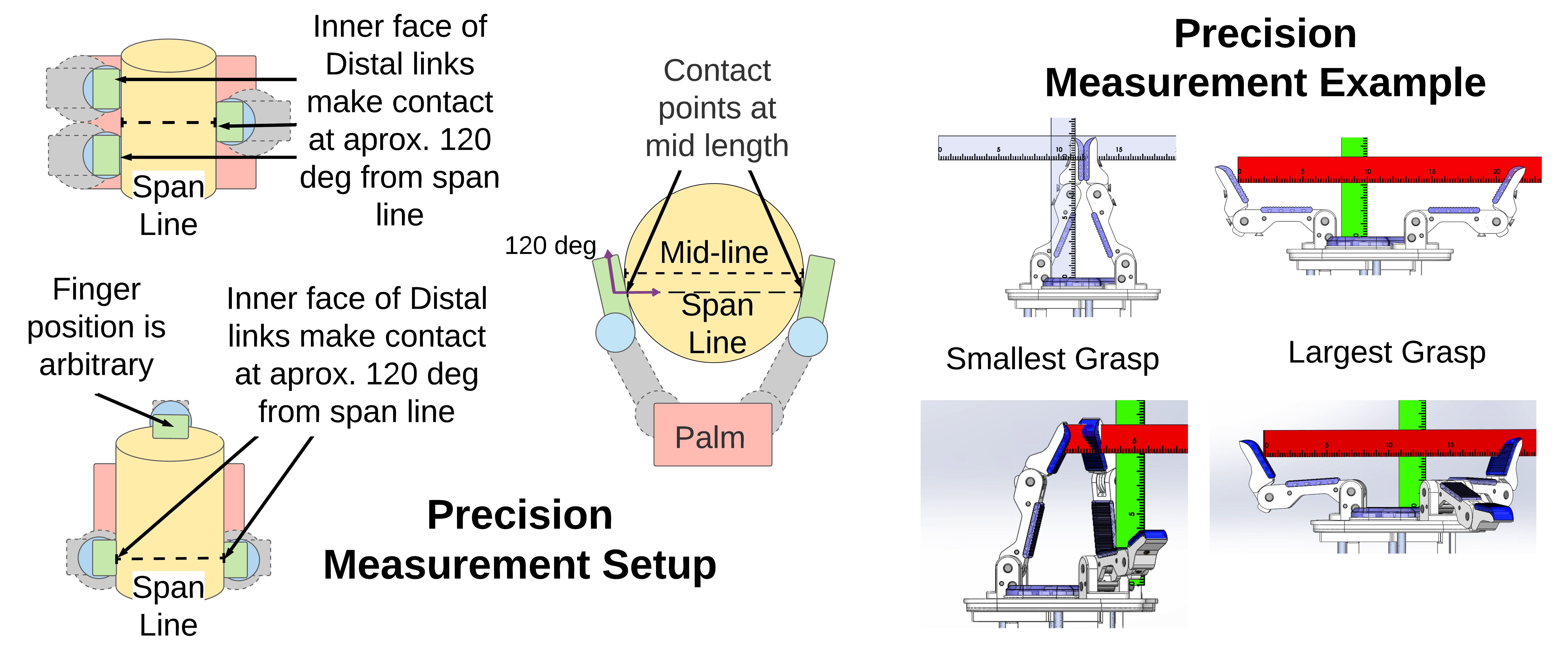}
    \caption{\textit{Left:} The precision grasp measurement process for a primarily left-right hand configuration (top left) and 3-fingered hand-configuration (bottom left). Fingers are placed roughly at the span-line, and should meet the surface close to perpendicular. Cylinder is provided for reference and is not required for the measurements. \textit{Right:} An example of taking the measurements using a virtual ruler on a CAD model.
    \vspace{-0.3cm}}
    \label{fig:precision}
\end{figure*}
%%%%%%%%%%%%%%%%%%%%%%%%%%%%%%%%%%%%%%%%%%%%%%%%%%%%

We discuss prior related work in two related domains: Benchmarking robot hand capabilities and representing robot kinematic workspaces.

\subsection{Benchmarking the Capabilities of Robot Hands}
An open problem in robot grasping and manipulation is the current lack of a standard for characterizing robot hands. Mahler et al discuss this problem and solutions, highlighting the need for better reporting of testing procedures in robot grasping research~\cite{bib:guesteditorial}. They especially emphasize the need for reporting `object ranges' that a hand can grasp. To this end, they recommended using object sets (a good sample:~\cite{bib:ycb-objectset,bib:amazon-objectset,bib:acrv-objectset,bib:sensorized-objectset, bib:grocery-objectset}).

Other benchmarks exist, such as those proposed by NIST \cite{bib:nist-benchmark}, however they specifically characterize {\em how} a hand can grasp, not {\em what} they can grasp.

In this work, we propose moving away from using specific objects to measure workspace size, and instead reporting an object's size {\em relative to} the hand's dimensions~\cite{bib:robothand-survey}. This normalized value is an intuitive way to communicate the salient features of an object's size in relation to the hand in question. For existing object sets~\cite{bib:ycb-objectset} and benchmarks~\cite{bib:benchmarks} this provides a simple mechanism for characterizing an object as, e.g., ``big'' or ``small'' relative to the hand in question.

\subsection{Representing Kinematic Workspaces of Robot Hands}
Kinematic workspaces are commonly used in research to characterize a hand's workspace~\cite{bib:human-precision-workspace,bib:interactivity-of-fingers,bib:learning-kinematic-workspace} or for determining possible caging grasps~\cite{bib:3dcaging, bib:caging}. Kinematic workspaces can be incredibly complex and difficult to represent in a human-readable way. This results in a lack of standardization in how roboticists communicate the functional space of their hand design, making between-hand comparison challenging.
%leading researchers to put together unique sets of experiments to characterize robot hands.

Our approach differs from previous measures because it is intended to be human-readable, an approach inspired by push grasping characterization~\cite{bib:push-grasping}. %In this work, the authors construct a simple volume indicating where an object can be pushed into a hand. 
Our work generates a volume in a similar manner, approximating the volume of the object that a hand can grasp.

\section{Hand-measurement Benchmark}
\label{sec:methods}
\begin{figure*}[t]
    \centering
    \includegraphics[width=0.95\textwidth]{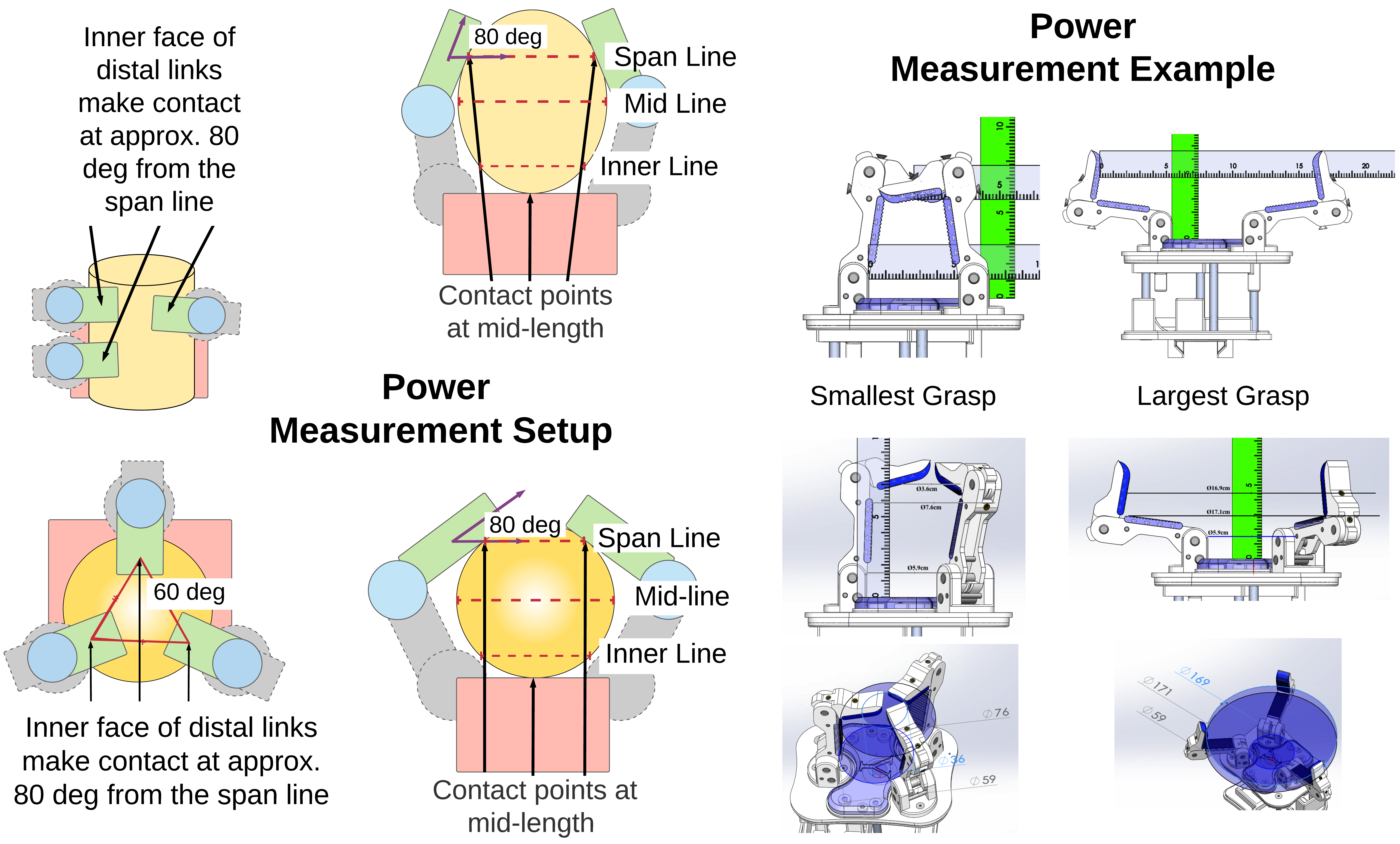}
    \caption{\textit{Left:} Details of the power grasp measurement process for grasping a cylinder (2D, top) and sphere (3D, bottom). %Additional information used to define the expected hand configuration and where to take the measurements at. 
    \textit{Right:} An example of taking the measurements using a virtual ruler on a CAD model (Model O).}
    \label{fig:power}
\end{figure*}

Our measurement approach is based on three concepts. First, we formally define a hand-based coordinate system based on functionality and hand orientation relative to a canonical object being grasped~\footnote{We use a cylinder and a sphere, since these objects are easy to define and the size of most real-world objects can be approximated with these shapes.} (Figure~\ref{fig:axes}). Second, we provide a quantitative definition for two common grasps,  Power and the Precision, with the goal of establishing the hand {\em configuration} required to take the measurements (Figures~\ref{fig:precision} and~\ref{fig:power}). Although these grasp definitions are usually described by how they restrict object motion, we provide additional positional and contact constraints to ensure that the measurements are repeatable. Third, we do not take measurements in a single hand finger pose, but instead at three poses in the kinematic workspace. These poses correspond to the largest, intermediate, and smallest grasp spaces. These measurements are parameterized based on the hand being ``open'' or ``closed'', and allow for a (rough) approximation of the usable volume in each coordinate axes. 

For a standard 2-fingered hand (or a 3-fingered hand with two fingers on a side) the most straightforward hand-based coordinate system consists of one vector pointing out of the palm (Depth), a second vector between the two opposite fingers (Span), and a third that points up out of the hand (Width). A Precision grasp is one which brings the fingers into contact on opposite sides of the object. A Power grasp is one where the fingers enclose the object (which is not possible with a traditional parallel jaw gripper). Our challenge is to extend these definitions to arbitrary hand configurations where there is {\em not} an obvious coordinate system, and where the fingers can enclose an object from arbitrary directions. For this reason, we turn to a more {\em functional} definition that asks the person measuring to envision how the hand might grasp the canonical object. Additionally, we provide guidelines on how the fingers should contact the object --- these are meant to ensure that, under reasonable friction and actuator torque assumptions, the hand would be able to grasp the object. These functional definitions and guidelines represent the trade-off between repeatability and generalizability --- we rely on the measurer to determine the actual hand configurations that meet the goals of the grasp type. 

%Firstly, we assume that there is a notion of a ``palm''. This palm defines the plane whose normal is our Depth axis. We define the coordinate system using a simple hand-object setup: grasping a cylinder placed on a table. 
%We use three hand-object configurations to define the coordinate systems, the grasps, and the measurements. 
%The first configuration is grasping a cylinder-like ellipsoid placed on a table. The second is holding an ellipsoidal object in the hand. The third is holding a long, thin ellipsoidal shape between the finger tips, parallel to the table. 
%We measure a hand's dimensions using two grasping scenarios. The first scenario is grasping the cylinder using a precision grasp, the second a power grasp.
%The setup and grasping scenarios are used to help ground the hand configurations and directions used in the measurements. By defining the {\em intent} of the grasp --- and letting the human decide what actual hand configuration meets the spirit of that grasp --- we introduce flexibility without needing to make complex constraints on the actual contact points and joint angles.

We next provide the functional definitions of the hand-based coordinate system (Section~\ref{sec:axes}), measurments that are independent of the grasp configuration (Section~\ref{sec:onetime}), the Precision and Power grasp configurations  (Sections~\ref{sec:precision} and~\ref{sec:power}), and a discussion of the implementation details we use to ensure the measurements are reproducible (Section~\ref{sec:accountable}).  Section~\ref{sec:usability} defines how to calculate object sizes relative to the hand measurements. Examples of the measurements are in Figures~\ref{fig:ConfA_Data} and~\ref{fig:ConfB_Data}.

\subsection{The Hand-based Axes: Span, Depth, and Width}
\label{sec:axes}
We define the hand-based coordinate using a simple grasping scenario: The measurer should position the hand so that it can grasp a cylinder (resting on a table) from the side. The axes are defined based on this measurer-selected hand pose, as shown in Figure~\ref{fig:axes}. The intention of this pose is to establish a clear left-right closing direction. While no actual cylinder is required, a physical cylinder-shaped object can help with visualizing the best way to orient the hand to achieve this grasp. 

%%%%%%%%%%%%%%%%%%%%%%%%%%%%%%%%%%%%%%%%%%%%%%%%%%%%
%%%%%%%%%%%%%%%%%%%%%%%%%%%%%%%%%%%%%%%%%%%%%%%%%%%%

\noindent \textbf{\underline{Span}}:
Span is the axis parallel to both the reference cylinder's faces and the palm's normal --- i.e., a line passing through the middle of the cylinder parallel to the table. Moving in the span axis is akin to moving the reference cylinder between the fingers across the table surface. This is used to specify the size of an object that can fit inside the hand in the direction of span. 

\noindent  \textbf{\underline{Depth}}:
Depth is the axis orthogonal to the plane of the palm. Moving in the depth axis is akin to moving the reference cylinder closer to, or farther from, the palm. Span and Depth are coupled quantities. If the opposing fingers are spaced further apart to increase span, the depth may decrease, and vice versa. 

\noindent \textbf{\underline{Width}}:
Width is the axis orthogonal to the reference cylinder's end-caps (up and down). Moving in the width axis would be akin to lifting the cylinder straight up off the table. The width axis provides a measure for the heights of objects that can be grasped by a hand. For most hands, the width measurements remain constant throughout actuation (see following section).

\subsection{One-time Measurements}
\label{sec:onetime}
In this section we describe measurements that are standalone measurements and are constant between Precision and Power grasp measurements. These are: Absolute Maximum Span and the Minimum and Maximum Width measurements. The Absolute Maximum Span is used to normalize span grasping regions relative to the physical size of the hand. The Minimum Width measures the shortest object the hand could pick up off of the table. The Maximum Width is the overall width of the hand.

The Absolute Maximum Span measures the distance between the ends of the furthest distals, without taking grasping into account. For this measurement, open the hand as far as it will go (actuated or not).  This is the measurement used in the push-grasping work~\cite{bib:push-grasping}. 

It is important to consider finger opposition when measuring width, as is the case for the asymmetrical hand design examples shown in Figure~\ref{fig:axes}. Minimum Width is the distance from the table top to the center of the two (or more) opposing fingers that would grasp the object. Minimum Width incorporates two fingers because the hand would fail to grasp a shorter object without the opposing finger.

Maximum Width measures the height of the hand resting on the table, {\em not} the tallest object, since that could (potentially) be infinite. To record whether or not the {\em object} height is infinite (i.e., not constrained), we add a `+' after the Maximum Width measurement.

\subsection{Precision Grasp Measurements}
\label{sec:precision}

Precision grasp measurements define how ``wide'' of an object can be grasped, and how that changes as the fingers close. We approximate this space by taking span and depth measurements at a minimum of three finger configurations, based on span: 1) maximum span, 2) minimum span, and 3+) at least one intermediate grasp between those poses. 

We define the Precision grasp as a stable grasp at the centerline of the object with contact points at the center of the distal link or at the fingertip. The measurer can choose whichever contact points give the maximum width, but this choice must remain the same for all of the measurements. Exactly how the fingers contact the object can vary widely; we chose as a guideline that the contact surfaces be oriented 30\degree outward (120\degree total) relative to the face of the palm, as illustrated on the left of Figure~\ref{fig:precision}. 
The intent of this angle constraint is to rule out grasps where the fingers contact the object but there is insufficient frictional force to keep the object from ``popping'' out of the grasp.

The corresponding Depth measurement for each Span measurement is the distance from the middle of the Span line (connecting the distals) to the palm (`Span Line' in Figure~\ref{fig:precision}). The Span line should be placed at the center of the distal links on opposing fingers. If there is more than one finger on each side, then chose the pair of fingers that would be used to grasp the cylinder.

\subsection{Power Grasp Measurements}
\label{sec:power}
Power grasp measurements define the volume of an object that a hand can power grasp, which we  again measure at three different finger poses. We approximate this volume at each finger pose by measuring three cross-sectional areas at a minimum of three depth locations inside the grasp: 1) largest possible power grasp volume, 2) smallest possible power grasp (a closed hand), and 3+) at least one intermediate grasp. 

We support two variations of the Power grasp, a spherical one that fully encloses the object from three (or more) directions, and a cylindrical grasp that encloses a (2D) circular area. Hands may be capable of none, one, or both grasps. The measurement of the Span-Depth pairs is different for the cylindrical and spherical power grasps, although the general approach is the same. For both cases, the fingers must contact the object {\em past} the centerline and at an angle of at least 80\degree relative to the Span line connecting the mid-points.

For the cylindrical volume measurements, the fingers are placed in the enclosing configuration and the Span-Depth pairs measured at the Inner, Mid, and Span lines since the fingers often ``bulge'' out (Top middle, Figure~\ref{fig:power}). For the spherical volume measurements, Width is constrained in addition to Depth. Rather than try to specify a width in the Span direction only, we chose the smallest direction in the Span-Width {\em plane}. Specifically, we measure  the diameter at the base of the fingers, the widest point, and the distal contact points.  

\begin{figure*}[t]
    \centering
    \includegraphics[width=0.975\linewidth]{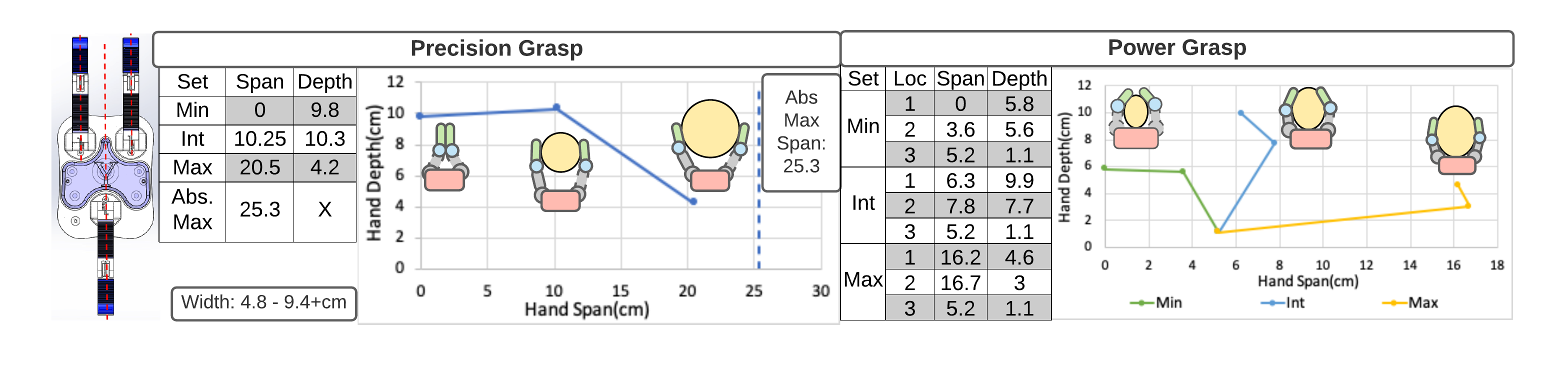}
    \caption{\textit{Left:} Yale Openhand Model O in a cylindrical configuration (fingers together). \textit{Middle:} Measurements and plot for precision cylindrical grasp. \textit{Right:} Measurements and plot for power cylindrical grasp. \vspace{-0.5cm}}
    \label{fig:ConfA_Data}
\end{figure*}

\begin{figure*}[t]
    \centering
    \includegraphics[width=0.975\linewidth]{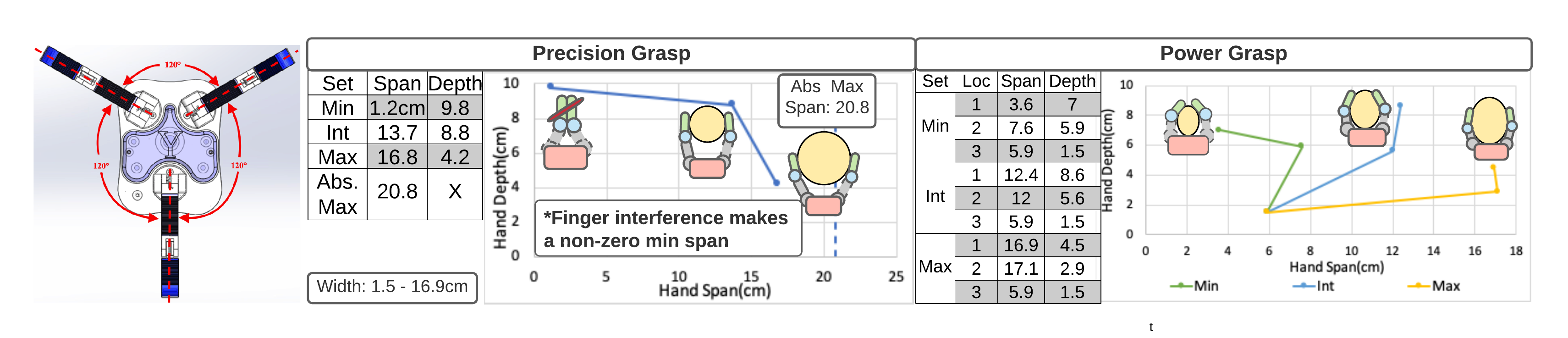}
    \caption{\textit{Left:} Yale Openhand Model O in a spherical  configuration (fingers 120\degree apart). \textit{Middle:} Measurements and plot for cylindrical precision grasp. \textit{Right:} Measurements and plot for spherical power grasp..
    \vspace{-0.5cm} }
    \label{fig:ConfB_Data}
\end{figure*}

\subsection{Using the Measurements}
\label{sec:usability}
As previously stated, we can use the hand measurements to a) normalize object size relative to the hand and b) represent an object-centric approximation of the hand's kinematic space.

We normalize object size with the values measured at the maximum $M$ and minimum $m$ span for precision grasps. When defining small, medium, and large objects we propose the following size ranges: 0-30\%, 31-69\%, and 70+\% of a hand's span $m + \%(M-m)$. We also define default values for small, medium, and large objects as: 25\%, 50\%, and 75\%. 

We represent our object-centric, approximated kinematic hand space by plotting span against depth (see Figures \ref{fig:ConfA_Data} \& \ref{fig:ConfB_Data}). The values in these plots can be interpolated in order to calculate a continuous approximation of the hand volume as the hand closes. 

We demonstrate using these measurements to classify a selection of YCB objects~\cite{bib:ycb-objectset} (see Figures~\ref{fig:ycbs} \& ~\ref{fig:rel-obj}). Note that, while our measurements can determine if an object would \textit{fit} in a hand, our measurements do not guarantee that the hand can successfully \textit{grasp} that object.

\subsection{Repeatable Measurements}
\label{sec:accountable}
Specifics and detailed instructions are provided in our github repository \cite{bib:github}. We include here a brief discussion on how to make the measurements repeatable and replicatable. First, we recommend taking pictures of each finger pose setup from two angles with a grid in the background. Second, if a kinematic CAD model exists then one can alternatively measure with virtual rulers in simulation (see Figures~\ref{fig:precision} and~\ref{fig:power}).

In practice it may be difficult to move hands into the desired positions, and keep them there, for measurement. This is especially true for underactuated hands performing precision and power grasps. In this case we suggest using an actual object to keep the hand in position.

\section{Measurement Examples}
\label{sec:results}

We demonstrate the resulting measurements for two finger configurations (cylindrical and spherical) of the Openhand Model O hand~\cite{bib:yale-openhand} (shown in Figures~\ref{fig:ConfA_Data} and~\ref{fig:ConfB_Data}).  We also include a comparison of relative object sizes using a selection of YCB objects for the Openhand configurations and five other hands, including the Openhand T-42, Robotiq 2F-85, Kinova Jaco 2 3 finger, one finger configuration of Barrett hand, and the right hand of the first author. Detailed results for the remaining five hands were left out for sake of space but are included on Github~\cite{bib:github}. Three hands were measured in collaboration with the NERVE lab (UMO).
%More hands will be added thanks to a collaboration with the NERVE lab (UML).

%To test the proposed measurement instructions researchers at the NERVE lab (UML) measured three hands: Yale Openhand T42~\cite{bib:yale-openhand}, Jaco2 3-finger~\cite{bib:Jaco2}, and Robotiq 2F-85~\cite{bib:robotiq} (also documented on GitHub).

The precision grasp plots in~\ref{fig:ConfA_Data} and~\ref{fig:ConfB_Data} illustrates the differences in the grasping space of each configuration. The area under each trend line represents the volume area of graspable object size, assuming adequate width. The spherical configuration has a minimum object size due to a non-zero minimum span that is not present on the cylindrical configuration. This lower bound is present because the distal links interfere with each other --- effectively blocking the fingers from closing further. This makes the cylindrical configuration more suited to grasping smaller objects.
%It is evident that spherical configuration has limitations in grasping smaller objects while a cylindrical configuration can theoretically grasp something infinitesimally thin, making it more suitable for tasks involving smaller objects.

%An general relationship between span and depth can be observed from the plots in Figure~\ref{fig:ConfA_Data} and Figure~\ref{fig:ConfB_Data}: as span decreases the depth increases past the mid point (at the peak) because the distal link rotates inwards (negative angle) to the palm, . The larger negative span-depth slope on the precision grasp in Figure~\ref{fig:ConfB_Data} is due to the distal links being limited to 120deg relative to the palm. As the fingers move inwards, it straightens making the angle between proximal and distal links zero and increases depth changes relative to span.

The power grasp plots in~\ref{fig:ConfA_Data} and~\ref{fig:ConfB_Data} illustrate the region of graspable space and how it changes with a hand's actuation. Assuming adequate width, one can fit an object's dimensions to these volumes to see whether an object will fit and approximate how much the hand will close before contact. The distal link interference is also present in the power grasp measurements for the spherical configuration. The cylindrical configuration --- due to the fact that the fingers can interlace --- is able to power grasp smaller objects than the spherical configuration. 

%The cylindrical configuration is more commonly featured in grasping research. Our findings validate this by showing that the cylindrical configuration is more well rounded at grasping --- whereas the spherical configuration is optimized for grasping spheres.

\begin{figure}
    \centering
    \includegraphics[width=1\linewidth]{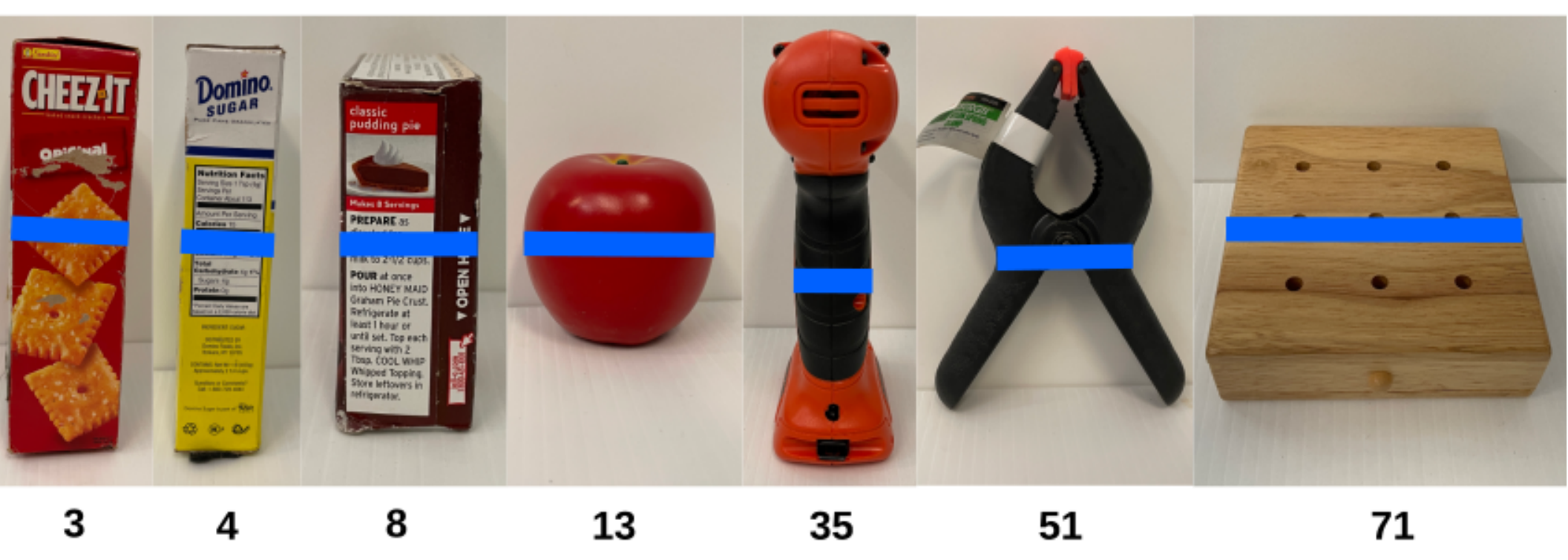}
    \caption{YCB objects, with identification numbers, used for relative object size analysis. Measurements taken at the blue line. \vspace{-0.3cm}}
    \label{fig:ycbs}
\end{figure}

Seven selected YCB objects were placed on a table and measured at the most likely grasp point (blue lines in Figure~\ref{fig:ycbs}). These sizes were then compared to the grasp size ranges for a precision grasp (Section~\ref{sec:usability}) for each hand and labeled as Small, Medium, Large, or Too large (Figure \ref{fig:rel-obj}). One interesting thing to note is that the {\em pattern} of sizes is different for each hand. This is because of hand size or due to the {\em actuation} of the hand. %the hand Span and Depth do not change in the same way as the hand closes.

\section{Discussion}
\label{sec:discussion}

\begin{figure}
    \centering
    \includegraphics[width=1\linewidth]{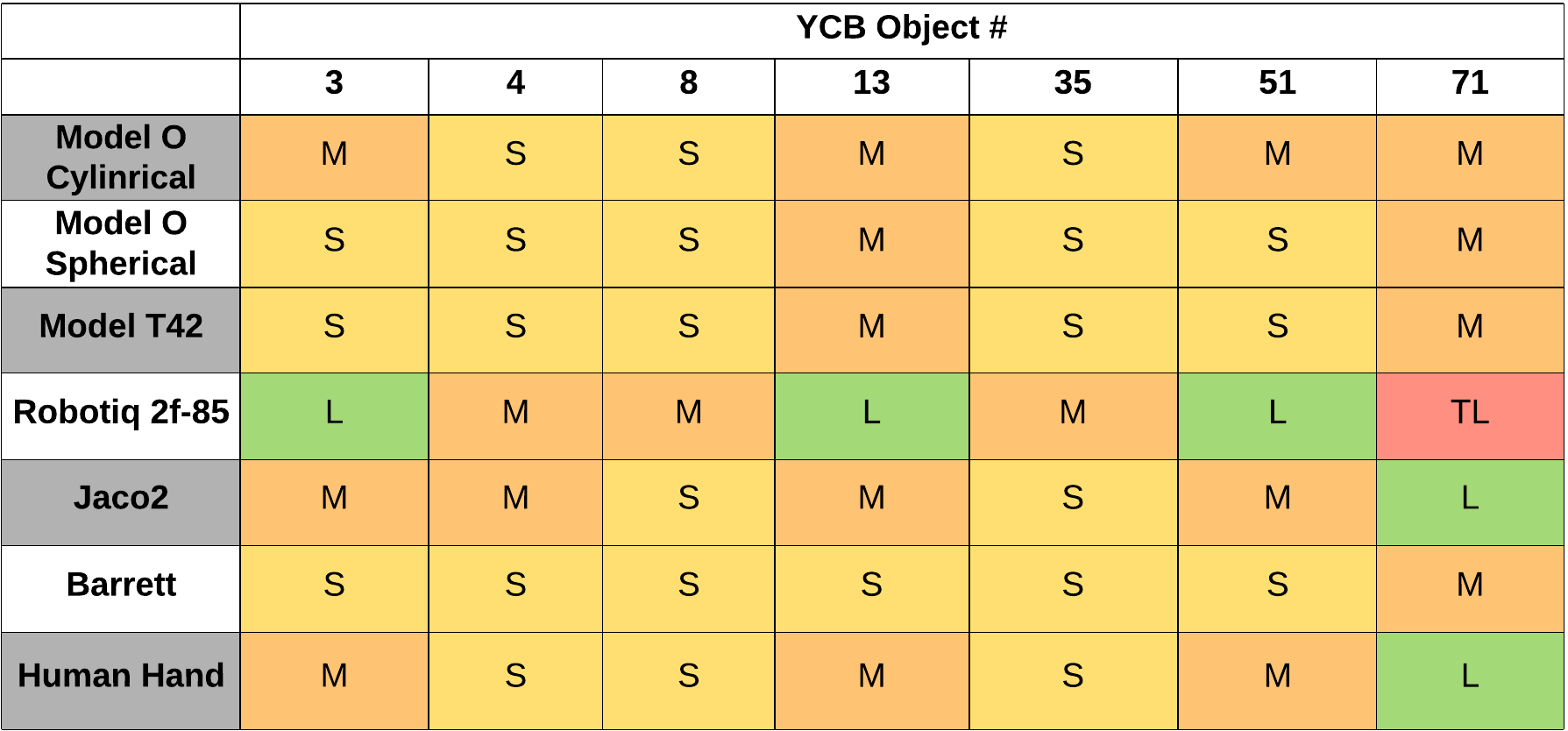}
    \caption{Each object's relative size with respect to each robot hand: small, medium, large, or too large. \vspace{-0.5cm}}
    \label{fig:rel-obj}
\end{figure}

Our results demonstrate how relative object size can be used to communicate an approximate kinematic workspace for a hand (Figures \ref{fig:ConfA_Data} \& \ref{fig:ConfB_Data}). Further, we show how relative object size can be used to effectively describe a grasping scenario, not necessarily predicting success, but effectively describing (for easy visualization) how the scenario will differ between hands for the same object (see Figure \ref{fig:rel-obj}).

When used in combination with an object set, these measures help clarify how well-suited a hand is for grasping those objects (or if it can grasp the object at all). For example, if all of the objects are categorized as small then  it \textit{suggests} that these objects will be difficult for that hand to grasp. This is not only useful for measuring graspable objects, but can be used to compare in-hand object \textit{translations} meaningfully between hand designs.

\subsection{Limitations and Suggested Improvements}
It is difficult to generalize the measurement axes to all types of fingered hands because of how varied and unique hand designs can be. By tying measurements to functional tasks we provide a method for characterizing a common subset of the hand's kinematic space --- however this is not a complete characterization of the space. We left the choice of how close an approximation to make (aka how many intermediate measurements to make) to those measuring the hands in order to accommodate a wider range of use cases. Also, using human judgment (and physical rulers) means that an element of error or noise will always exist in these measurements.

%There may be more than one ``solution'' that satisfies the functional task requirements. In this case, we rely on the configuration data captured during the measurement process (Section~\ref{sec:accountable}) to document the different solutions. 

%One option would be to perform these measurements automatically, for example using computer vision or simulations, with guidance from the human on pose configurations. 

%We demonstrate measurements for only two relatively simple grasps. %Obviously, this does not begin to describe the kinematic work space of more complex hands. 
%Although we specify a measuring minimum of three finger poses for each grasp, this may not sufficiently describe the space.

%Additionally only measurements for three different finger poses are required, which may not be enough to catch all aspects of the finger's actuation. As duplicate submissions occur the measurements can be combined, mitigating this concern. 

\vspace{-0.2cm}
%\newpage
%\appendix
%\label{sec:appendix}
%\input{06appendix2.tex}

%\newpage
\section*{Acknowledgment}
This work supported in part by NSF grants CCRI 1925715 and RI 1911050. We also thank Adam Norton and Brian Flynn at UML for profitable discussions.

\ifCLASSOPTIONcaptionsoff
  \newpage
\fi

\bibliographystyle{IEEEtran}
\bibliography{references}

\begin{thebibliography}{10}
\providecommand{\url}[1]{#1}
\csname url@samestyle\endcsname
\providecommand{\newblock}{\relax}
\providecommand{\bibinfo}[2]{#2}
\providecommand{\BIBentrySTDinterwordspacing}{\spaceskip=0pt\relax}
\providecommand{\BIBentryALTinterwordstretchfactor}{4}
\providecommand{\BIBentryALTinterwordspacing}{\spaceskip=\fontdimen2\font plus
\BIBentryALTinterwordstretchfactor\fontdimen3\font minus
  \fontdimen4\font\relax}
\providecommand{\BIBforeignlanguage}[2]{{%
\expandafter\ifx\csname l@#1\endcsname\relax
\typeout{** WARNING: IEEEtran.bst: No hyphenation pattern has been}%
\typeout{** loaded for the language `#1'. Using the pattern for}%
\typeout{** the default language instead.}%
\else
\language=\csname l@#1\endcsname
\fi
#2}}
\providecommand{\BIBdecl}{\relax}
\BIBdecl

\bibitem{bib:scientific-method}
G.~Antonelli, ``Robotic research: are we applying the scientific method?''
  \emph{Frontiers in Robotics and AI}, vol.~2, p.~13, 2015.

\bibitem{bib:guesteditorial}
J.~Mahler, R.~Platt, A.~Rodriguez, M.~Ciocarlie, A.~Dollar, R.~Detry, M.~A.
  Roa, H.~Yanco, A.~Norton, J.~Falco \emph{et~al.}, ``Guest editorial open
  discussion of robot grasping benchmarks, protocols, and metrics,'' \emph{IEEE
  Trans. on Automation Science and Engineering}, vol.~15, no.~4, pp.
  1440--1442, 2018.

\bibitem{bib:ycb-objectset}
B.~Calli, A.~Singh, A.~Walsman, S.~Srinivasa, P.~Abbeel, and A.~M. Dollar,
  ``The ycb object and model set: Towards common benchmarks for manipulation
  research,'' in \emph{IEEE ICAR}.\hskip 1em plus 0.5em minus 0.4em\relax IEEE,
  2015, pp. 510--517.

\bibitem{bib:sensorized-objectset}
G.~Gao, G.~Gorjup, R.~Yu, P.~Jarvis, and M.~Liarokapis, ``Modular, accessible,
  sensorized objects for evaluating the grasping and manipulation capabilities
  of grippers and hands,'' \emph{IEEE Robotics and Automation Letters}, vol.~5,
  no.~4, pp. 6105--6112, 2020.

\bibitem{bib:task-based}
V.~Ortenzi, M.~Controzzi, F.~Cini, J.~Leitner, M.~Bianchi, M.~A. Roa, and
  P.~Corke, ``Robotic manipulation and the role of the task in the metric of
  success,'' \emph{Nature Machine Intelligence}, vol.~1, no.~8, pp. 340--346,
  2019.

\bibitem{bib:grasp-taxonomy}
T.~Feix, R.~Pawlik, H.-B. Schmiedmayer, J.~Romero, and D.~Kragic, ``A
  comprehensive grasp taxonomy,'' in \emph{Robotics, science and systems:
  workshop on understanding the human hand for advancing robotic manipulation},
  vol.~2, no. 2.3.\hskip 1em plus 0.5em minus 0.4em\relax Seattle, WA, USA;,
  2009, pp. 2--3.

\bibitem{bib:yale-openhand}
R.~Ma and A.~Dollar, ``Yale openhand project: Optimizing open-source hand
  designs for ease of fabrication and adoption,'' \emph{IEEE Robotics \&
  Automation Magazine}, vol.~24, no.~1, pp. 32--40, 2017.

\bibitem{bib:github}
``Applied hand dimensions,''
  \url{https://github.com/OSUrobotics/Applied-Hand-Dimensions}, 2020.

\bibitem{bib:amazon-objectset}
N.~Correll, K.~E. Bekris, D.~Berenson, O.~Brock, A.~Causo, K.~Hauser, K.~Okada,
  A.~Rodriguez, J.~M. Romano, and P.~R. Wurman, ``Analysis and observations
  from the first amazon picking challenge,'' \emph{IEEE Trans. on Automation
  Science and Engineering}, vol.~15, no.~1, pp. 172--188, 2016.

\bibitem{bib:acrv-objectset}
J.~Leitner, A.~W. Tow, N.~S{\"u}nderhauf, J.~E. Dean, J.~W. Durham, M.~Cooper,
  M.~Eich, C.~Lehnert, R.~Mangels, C.~McCool \emph{et~al.}, ``The acrv picking
  benchmark: A robotic shelf picking benchmark to foster reproducible
  research,'' in \emph{IEEE ICRA}.\hskip 1em plus 0.5em minus 0.4em\relax IEEE,
  2017, pp. 4705--4712.

\bibitem{bib:grocery-objectset}
P.~Triantafyllou, H.~Mnyusiwalla, P.~Sotiropoulos, M.~A. Roa, D.~Russell, and
  G.~Deacon, ``A benchmarking framework for systematic evaluation of robotic
  pick-and-place systems in an industrial grocery setting,'' in \emph{IEEE
  ICRA}.\hskip 1em plus 0.5em minus 0.4em\relax IEEE, 2019, pp. 6692--6698.

\bibitem{bib:nist-benchmark}
J.~Falco, D.~Hemphill, K.~Kimble, E.~Messina, A.~Norton, R.~Ropelato, and
  H.~Yanco, ``Benchmarking protocols for evaluating grasp strength, grasp cycle
  time, finger strength, and finger repeatability of robot end-effectors,''
  \emph{IEEE Robotics and Automation Letters}, vol.~5, no.~2, pp. 644--651,
  2020.

\bibitem{bib:robothand-survey}
A.~Kothari, J.~Morrow, V.~Thrasher, K.~Engle, R.~Balasubramanian, and C.~Grimm,
  ``Grasping objects big and small: Human heuristics relating grasp-type and
  object size,'' in \emph{IEEE ICRA}.\hskip 1em plus 0.5em minus 0.4em\relax
  IEEE, 2018, pp. 1--6.

\bibitem{bib:benchmarks}
Y.~Bekiroglu, N.~Marturi, M.~A. Roa, K.~J.~M. Adjigble, T.~Pardi, C.~Grimm,
  R.~Balasubramanian, K.~Hang, and R.~Stolkin, ``Benchmarking protocol for
  grasp planning algorithms,'' \emph{IEEE Robotics and Automation Letters},
  vol.~5, no.~2, pp. 315--322, 2019.

\bibitem{bib:human-precision-workspace}
I.~M. Bullock, T.~Feix, and A.~M. Dollar, ``Workspace shape and characteristics
  for human two-and three-fingered precision manipulation,'' \emph{IEEE Trans.
  on Biomedical Engineering}, vol.~62, no.~9, pp. 2196--2207, 2015.

\bibitem{bib:interactivity-of-fingers}
W.~S. You, Y.~H. Lee, G.~Kang, H.~S. Oh, J.~K. Seo, and H.~R. Choi, ``Kinematic
  design optimization for anthropomorphic robot hand based on interactivity of
  fingers,'' \emph{Intelligent Service Robotics}, vol.~12, no.~2, pp. 197--208,
  2019.

\bibitem{bib:learning-kinematic-workspace}
L.~Jamone, L.~Natale, G.~Sandini, and A.~Takanishi, ``Interactive online
  learning of the kinematic workspace of a humanoid robot,'' in \emph{IEEE
  IROS}.\hskip 1em plus 0.5em minus 0.4em\relax IEEE, 2012, pp. 2606--2612.

\bibitem{bib:3dcaging}
S.~Makita and Y.~Maeda, ``3d multifingered caging: Basic formulation and
  planning,'' in \emph{IEEE IROS}.\hskip 1em plus 0.5em minus 0.4em\relax IEEE,
  2008, pp. 2697--2702.

\bibitem{bib:caging}
S.~Makita and W.~Wan, ``A survey of robotic caging and its applications,''
  \emph{Advanced Robotics}, vol.~31, no. 19-20, pp. 1071--1085, 2017.

\bibitem{bib:push-grasping}
M.~R. Dogar and S.~S. Srinivasa, ``Push-grasping with dexterous hands:
  Mechanics and a method,'' in \emph{IEEE IROS}.\hskip 1em plus 0.5em minus
  0.4em\relax IEEE, 2010, pp. 2123--2130.

\end{thebibliography}

\end{document}